# REAL-TIME IMPLEMENTATION OF ORDER-STATISTICS

# BASED DIRECTIONAL FILTERS

**M. Emre Celebi**

ecelebi@lsus.edu

Department of Computer Science, Louisiana State University in Shreveport

One University Place, Shreveport, LA 71115, USA (P: 318-795-4281; F: 318-795-2419)

# ABSTRACT

Vector filters based on order-statistics have proved successful in removing impulsive noise from color images while preserving edges and fine image details. Among these filters, the ones that involve the cosine distance function (directional filters) have particularly high computational requirements, which limits their use in time critical applications. In this paper, we introduce two methods to speed up these filters. Experiments on a diverse set of color images show that the proposed methods provide substantial computational gains without significant loss of accuracy.

# I. INTRODUCTION

Color images are often contaminated with noise, which is introduced during acquisition or transmission. In particular, the introduction of impulsive noise into an image not only lowers its perceptual quality, but also makes subsequent tasks such as edge detection and segmentation more difficult. Therefore, the removal of such noise is often an essential preprocessing step in many color image processing applications.

Numerous filters have been proposed for the removal of impulsive noise from color images [1][2][3]. Among these, nonlinear filters have proved successful in removing the noise while preserving the edges and fine details. The early approaches to nonlinear filtering of color images often involved the application of a scalar filter to each color channel independently. However, since separate processing ignores the inherent correlation between the color channels, these methods often introduce color artifacts to which the human visual system is very sensitive. Therefore, vector filtering techniques that treat the color image as a vector field and process color pixels as vectors are more appropriate. An important class of nonlinear vector filters is the one based on robust order-statistics with the vector median filter (*VMF*) [4], the basic vector directional filter (*BVDF*) [5], and the directional-distance filter (*DDF*) [6] being the most widely known examples. These filters involve reduced ordering [7] of a set of input vectors within a window to determine the output vector. The ordering is often achieved using a combination of two functions: Minkowski distance and cosine distance. The *VMF* and its derivatives use the former, whereas the *VDF* family (the *BVDF* and its derivatives) use the latter. The *DDF* family uses a combination of the two functions. In this paper, we refer to the filters that use the cosine distance function, i.e. the members of *VDF* and *DDF* families, as *directional filters*.

Several researchers have noted the high computational requirements of order-statistics based vector filters; however, relatively few studies have focused on alleviating this problem. Chaudhuri et al. [8] proposed an approximate $L_2$ (Euclidean) metric, which is calculated as a linear combination of the $L_1$ (City-block) and $L_\infty$ (Chessboard) metrics. Barni et al. [9] proposed a more accurate approximation for the $L_2$ metric, which is calculated as a weighted $L_1$ metric. Note that the utility of these approximations is limited to the *VMF* family. In a recent study [3], we compared 48 order-statistics based vector filters and concluded that some of the most effective filters are based on the cosine distance function. In fact, one of these filters, namely the adaptive center weighted directional distance filter (*ACWDDF*) [10], was shown to be the most effective filter. It was also shown that the directional filters are significantly slower than those based only on the Minkowski distance.

In this paper, we introduce techniques to speed up the order-statistics based directional filters. Experiments on a diverse set of color images show that presented methods achieve substantial computational gains without significant loss of accuracy. The rest of the paper is organized as follows. Section II introduces the notation and describes the techniques to speed up the cosine distance function. Section III presents the experimental results. Finally, Section IV gives the conclusions.

## II. PROPOSED METHODS

Consider an $M$ x $N$ RGB image $\mathbf{X}$ that represents a two-dimensional array of three-component vectors $\mathbf{x}(r,c) = [x_1(r,c), x_2(r,c), x_3(r,c)]$ occupying the spatial location $(r,c)$, with the row and column indices $r = 1,2,...,M$ and $c = 1,2,...,N$, respectively. In pixel $\mathbf{x}(r,$c$)$, the $x_k(r,c)$ values denote the red ($k = 1$), green ($k = 2$), and blue ($k = 3$) component. In order to isolate small image regions, each of which can be treated as stationary, an $\sqrt{n} \times \sqrt{n}$ supporting window $W(r,c)$ centered on pixel $\mathbf{x}(r,c)$ is used. The window slides over the entire image $\mathbf{X}$ and the procedure replaces the input vector $\mathbf{x}(r, c)$ with the output vector $\mathbf{y}(r,c) = F(W(r,$c$))$ of a filter function $F(.)$ that operates over the samples inside $W(r,c)$. Repeating the procedure for each pair $(r,c)$, with $r = 1,2,...,M$ and $c = 1,2,...,N$, produces output vectors $\mathbf{y}(r,c)$ of the $M$ x $N$ filtered image $\mathbf{Y}$. For notational simplicity, the vectors inside $W(r,c)$ are re-indexed as $W(r,c) = \{\mathbf{x}_i : i = 1,2,...,n\}$, as commonly seen in the related literature [1][2][3]. Thus, in the vector $\mathbf{x}_i = [x_{i1}, x_{i2}, x_{i3}]$ with components $x_{ik}$, the $i$ and $k$ indices denote the block location and color channel, respectively.

The *VMF* family orders the input vectors in a window according to their relative magnitude differences using the Minkowski distance function. For example, the output of the *VMF*, the most well-known member of its class, is given by the lowest ranked input vector:

$$\mathbf{y}(r,c) = \operatorname*{argmin}_{\mathbf{x}_i \in W(r,c)} \left( \sum_{j=1}^{n} L_p(\mathbf{x}_i, \mathbf{x}_j) \right)$$

$$L_p(\mathbf{x}_i, \mathbf{x}_j) = \left( \sum_{k=1}^{3} \left| x_{ik} - x_{jk} \right|^p \right)^{1/p} \qquad (1)$$

where $L_p$ denotes the Minkowski distance.

The *VDF* family operates on the direction of the color vectors with the aim of eliminating vectors with atypical directions. The input vectors in a window are ordered according to their angular differences using the cosine distance function. For example, the output of the *BVDF*, the most well-known member of its class, is the input vector in the window whose direction is the maximum likelihood estimate of the input vector directions [11]:

$$\mathbf{y}(r,c) = \underset{\mathbf{x}_i \in W(r,c)}{\operatorname{argmin}} \left( \sum_{j=1}^{n} A(\mathbf{x}_i, \mathbf{x}_j) \right)$$

$$A(\mathbf{x}_i, \mathbf{x}_j) = \operatorname{acos} \left( \frac{\sum_{k=1}^{3} x_{ik} x_{jk}}{\|\mathbf{x}_i\| \|\mathbf{x}_j\|} \right) \quad (2)$$

where $A(\mathbf{x}_i, \mathbf{x}_j)$ denotes the angle between the two vectors $\mathbf{x}_i$ and $\mathbf{x}_j$, and $\|\cdot\|$ is the $L_2$ norm.

The *DDF* family combines the *VMF* and *VDF* families by simultaneously minimizing their ordering functions. The output of the *DDF*, the most well-known member of its class, is given by:

$$\mathbf{y}(r,c) = \underset{\mathbf{x}_i \in W(r,c)}{\operatorname{argmin}} \left( \left( \sum_{j=1}^{n} A(\mathbf{x}_i, \mathbf{x}_j) \right) \cdot \left( \sum_{j=1}^{n} L_p(\mathbf{x}_i, \mathbf{x}_j) \right) \right) \quad (3)$$

As mentioned in Section I, the *VDF* and *DDF* family members have much higher computational requirements than the *VMF* family members. This is due to the computationally expensive cosine distance function $A(\cdot, \cdot)$ used in (2) and (3). For example, on a typical 1024 x 1024 image, the *VMF* takes about 1.39 seconds, while the *BVDF* and *DDF* take approximately 27.9 and 28.5 seconds♣, respectively. In the following subsections, we introduce techniques to speed up the directional filters, i.e. the members of the *VDF* and *DDF* families.

## 2.1 Method 1

This method involves approximating the inverse cosine (*ACOS*) function in $A(\cdot, \cdot)$ using a minimax polynomial [12] of degree $q$:

---

♣ Programming language: C, Compiler: gcc 3.4.4, CPU: Intel Pentium D 2.66Ghz

$$P_q([a,b]) = \{a_0 + a_1 z + \ldots + a_q z^q : z \in [a,b], a_i \in \mathbb{R}, i = 0,1,\ldots q\} \quad (4)$$

The *ACOS* function takes arguments from the interval [0,1]. Unfortunately, approximating *ACOS* over this interval is not easy because of its behavior near 1 (see Fig. 1a).

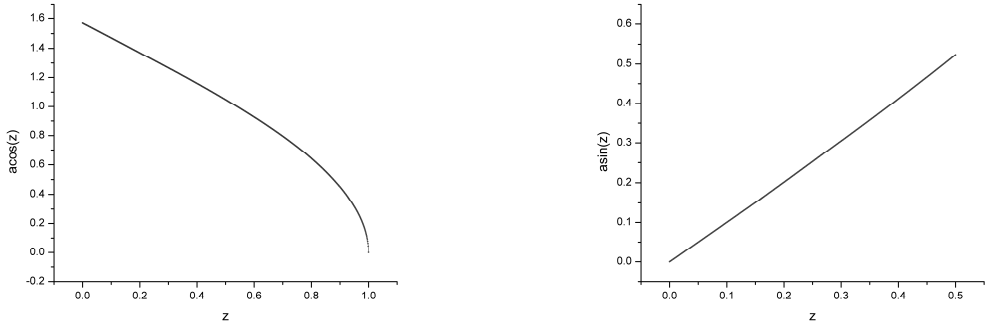

**Figure 1.** (a) *ACOS* function in the interval [0,1]  (b) *ASIN* function in the interval [0,0.5]

This can be circumvented using a numerically more stable identity for $z \geq 0.5$:

$$\text{acos}(z) = 2 \cdot \text{asin}\left(\sqrt{0.5(1-z)}\right) \quad (5)$$

where the inverse sine (*ASIN*) function receives its arguments from the interval [0,0.5] (see Fig. 1b). Instead of plugging the value of $\sqrt{0.5(1-z)}$ into a minimax approximation for the *ASIN* function and then multiplying the result by 2, two multiplication operations can be avoided if the following function is approximated:

$$\tau = \sqrt{1-z}$$
$$\text{acos}(z) = 2 \cdot \text{asin}\left(\tau / \sqrt{2}\right) \quad (6)$$

where the argument $\tau$ falls into the interval $[0, 1/\sqrt{2}]$.

Table 1 lists the coefficients of the minimax polynomials of various degrees for the approximation of the *ASIN* and *ACOS* functions. Since both functions exhibit strong linearity in their respective intervals, they can be accurately approximated by polynomials, as indicated by the small error values ( $\varepsilon$ ) in the table.

**Table 1.** Minimax polynomials for the *ASIN* and *ACOS* functions

| | $q$ | $\varepsilon$ | $a_0$ | $a_1$ | $a_2$ | $a_3$ | $a_4$ |
|---|---|---|---|---|---|---|---|
| A S I N | 2 | 1.830987519e-03 | 1.829125e-03 | 1.371117 | 1.480266e-01 | | |
| | 3 | 1.358426903e-04 | -1.358425e-04 | 1.419488 | -3.090315e-02 | 1.666491e-01 | |
| | 4 | 2.097813673e-05 | 2.097797e-05 | 1.412840 | 1.429881e-02 | 6.704361e-02 | 6.909677e-02 |
| A C O S | 2 | 9.154936808e-04 | 1.569882 | -9.695260e-01 | -1.480266e-01 | | |
| | 3 | 6.792158693e-05 | 1.570864 | -1.003730 | 3.090318e-02 | -2.356775e-01 | |
| | 4 | 1.048948667e-05 | 1.570786 | -9.990285e-01 | -1.429899e-02 | -9.481335e-02 | -1.381942e-01 |

## 2.2 Method 2

This method involves the substitution of the function $A(\cdot,\cdot)$ with a computationally cheaper function $B(\cdot,\cdot)$:

$$B(\mathbf{x}_i, \mathbf{x}_j) = L_p(\mathbf{x}_i', \mathbf{x}_j') = \left( \left| r_i - r_j \right|^p + \left| g_i - g_j \right|^p + \left| b_i - b_j \right|^p \right)^{1/p}$$

$$\mathbf{x}_i' = [r_i, g_i, b_i] = \left[ \frac{x_{i1}}{x_{i1} + x_{i2} + x_{i3}}, \frac{x_{i2}}{x_{i1} + x_{i2} + x_{i3}}, \frac{x_{i3}}{x_{i1} + x_{i2} + x_{i3}} \right] \qquad (7)$$

Here, $B(\cdot,\cdot)$ is a Minkowski distance function in the chromaticity coordinate space ($rgb$) [1].

# III. EXPERIMENTAL RESULTS

In this section, we evaluate the performance of the proposed methods on a set of test images commonly used in the color image filtering literature. Figure 2 shows representative images from this set. In the experiments, the filtering window was set to 3 x 3 and whenever the Minkowski distance is involved the $L_2$-norm was used as commonly seen in the related literature [1][2][3].

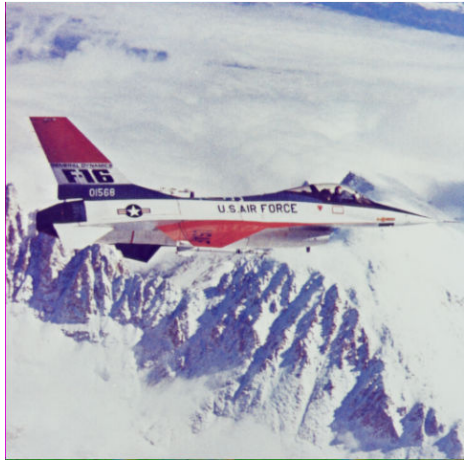

a) Airplane

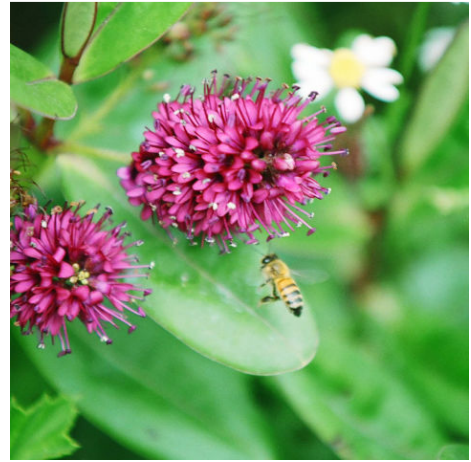

b) Flowerbee

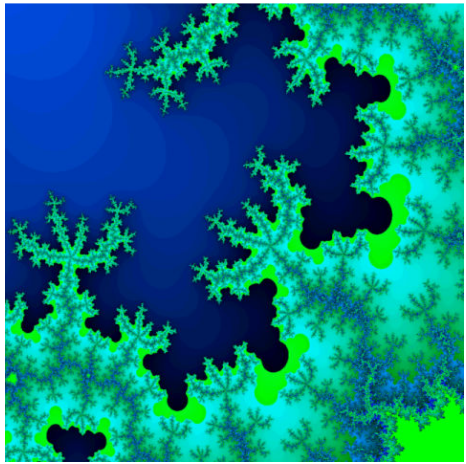

c) Fractal

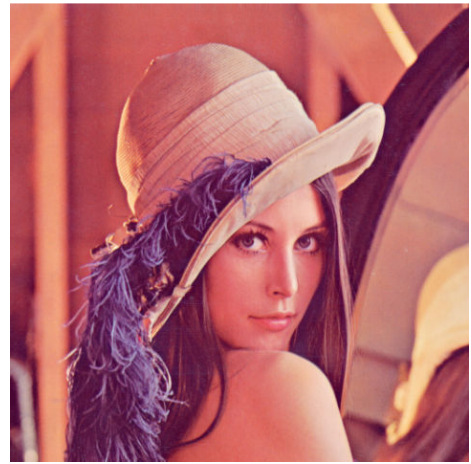

d) Lenna

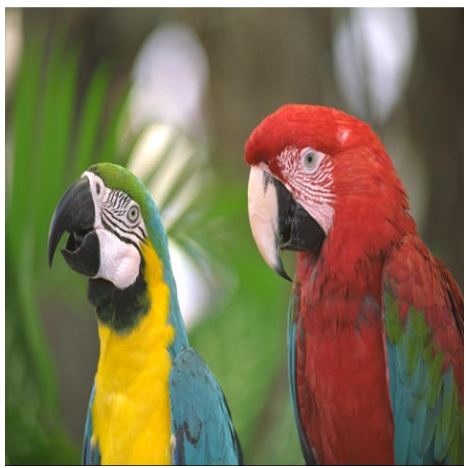

e) Parrots

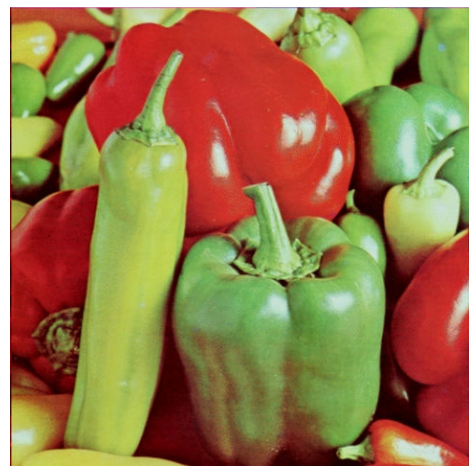

f) Peppers

**Figure 2.** Representative images from the test set

The corruption in the test images was simulated by the widely used correlated impulsive noise model [13]:

$$\mathbf{x} = \begin{cases} \mathbf{o} & \text{with probability } 1-\varphi, \\ \{r_1, o_2, o_3\} & \text{with probability } \varphi_1 \cdot \varphi, \\ \{o_1, r_2, o_3\} & \text{with probability } \varphi_2 \cdot \varphi, \\ \{o_1, o_2, r_3\} & \text{with probability } \varphi_3 \cdot \varphi, \\ \{r_1, r_2, r_3\} & \text{with probability } \left(1 - (\varphi_1 + \varphi_2 + \varphi_3)\right) \cdot \varphi \end{cases} \qquad (8)$$

where $\mathbf{o} = \{o_1, o_2, o_3\}$ and $\mathbf{x} = \{x_1, x_2, x_3\}$ represent the original and noisy color vectors, respectively, $\mathbf{r} = \{r_1, r_2, r_3\}$ is a random vector that represents the impulsive noise, $\varphi$ is the sample corruption probability, and $\varphi_1$, $\varphi_2$, and $\varphi_3$ are the corruption probabilities for the red, green, and blue channels, respectively. In the simulations, the channel corruption probabilities were set to 0.25.

Filtering performance was evaluated using three effectiveness and one efficiency criteria. The effectiveness criteria were the mean absolute error (*MAE*) [1], peak signal-to-noise ratio (*PSNR*) [1], and normalized color distance (*NCD*)♠ [1], which correspond, respectively, to signal detail preservation, noise suppression, and color-information preservation. Note that for the *MAE* and *NCD* measures lower values are better, whereas for the *PSNR* higher values are better. The efficiency of a filter was measured by the execution time in seconds♦.

### 3.1 Experiment 1

Tables 2-3 show the filtering results for the original *BVDF* (denoted by *BVDF*), method 1 (denoted by *BVDF$_{minimax}$*), and method 2 (denoted by *BVDF$_{rgb}$*). It can be seen that, in terms of filtering effectiveness, the *BVDF$_{minimax}$* and *BVDF$_{rgb}$* consistently performed similar to the *BVDF*. In fact, in some cases, they performed slightly better than the *BVDF*. With respect to the execution speed, on the average, the *BVDF$_{minimax}$* and *BVDF$_{rgb}$* were respectively 13.74 and 31.50 times faster than the *BVDF*.

---

♠ In the tables, the *NCD* values are multiplied by 1000.

♦ The results are the average of 10 runs.

**Table 2.** *BVDF* filtering results at 10% noise level

| Filter | Airplane (512 x 512 pixels) | | | | Flowerbee (480 x 480 pixels) | | | |
|---|---|---|---|---|---|---|---|---|
| | MAE | PSNR | NCD | TIME | MAE | PSNR | NCD | TIME |
| none | 6.400 | 17.865 | 82.798 | 0.000 | 6.338 | 18.204 | 89.361 | 0.000 |
| *BVDF* | 3.273 | 31.506 | 24.648 | 10.544 | 2.801 | 31.327 | 20.092 | 8.923 |
| $BVDF_{minimax}$ | 3.278 | 31.508 | 24.682 | 0.694 | 2.810 | 31.322 | 20.144 | 0.608 |
| $BVDF_{rgb}$ | 3.276 | 31.496 | 24.669 | 0.306 | 2.814 | 31.294 | 20.207 | 0.267 |
| | Fractal (640 x 480 pixels) | | | | Lenna (512 x 512 pixels) | | | |
| none | 6.361 | 17.141 | 102.837 | 0.000 | 6.369 | 18.195 | 102.593 | 0.000 |
| *BVDF* | 5.554 | 24.100 | 77.814 | 9.094 | 3.929 | 31.747 | 44.508 | 10.583 |
| $BVDF_{minimax}$ | 5.565 | 24.096 | 77.939 | 0.821 | 3.929 | 31.747 | 44.511 | 0.697 |
| $BVDF_{rgb}$ | 5.577 | 24.071 | 78.168 | 0.344 | 3.933 | 31.732 | 44.586 | 0.302 |
| | Parrots (1536 x 768 pixels) | | | | Peppers (512 x 480 pixels) | | | |
| none | 6.348 | 18.123 | 117.100 | 0.000 | 6.371 | 17.964 | 95.057 | 0.000 |
| *BVDF* | 0.916 | 38.993 | 7.378 | 50.495 | 2.263 | 32.538 | 18.876 | 9.498 |
| $BVDF_{minimax}$ | 0.969 | 39.054 | 7.708 | 4.147 | 2.277 | 32.391 | 19.035 | 0.658 |
| $BVDF_{rgb}$ | 0.921 | 38.925 | 7.414 | 1.786 | 2.271 | 32.571 | 18.983 | 0.294 |

**Table 3.** *BVDF* filtering results at 15% noise level

| Filter | Airplane (512 x 512 pixels) | | | | Flowerbee (480 x 480 pixels) | | | |
|---|---|---|---|---|---|---|---|---|
| | MAE | PSNR | NCD | TIME | MAE | PSNR | NCD | TIME |
| none | 9.553 | 16.131 | 123.414 | 0.000 | 9.537 | 16.418 | 134.211 | 0.000 |
| *BVDF* | 3.439 | 30.878 | 25.961 | 10.481 | 3.047 | 30.639 | 22.121 | 8.863 |
| $BVDF_{minimax}$ | 3.445 | 30.876 | 25.997 | 0.694 | 3.054 | 30.637 | 22.163 | 0.608 |
| $BVDF_{rgb}$ | 3.443 | 30.862 | 25.978 | 0.309 | 3.057 | 30.626 | 22.223 | 0.267 |
| | Fractal (640 x 480 pixels) | | | | Lenna (512 x 512 pixels) | | | |
| none | 9.560 | 15.373 | 153.478 | 0.000 | 9.560 | 16.429 | 153.113 | 0.000 |
| *BVDF* | 5.726 | 23.756 | 80.657 | 9.025 | 4.090 | 31.264 | 46.161 | 10.491 |
| $BVDF_{minimax}$ | 5.741 | 23.741 | 80.874 | 0.824 | 4.090 | 31.264 | 46.164 | 0.703 |
| $BVDF_{rgb}$ | 5.767 | 23.690 | 81.175 | 0.350 | 4.093 | 31.248 | 46.215 | 0.309 |
| | Parrots (1536 x 768 pixels) | | | | Peppers (512 x 480 pixels) | | | |
| none | 9.514 | 16.363 | 175.817 | 0.000 | 9.576 | 16.199 | 142.114 | 0.000 |
| *BVDF* | 1.003 | 37.935 | 8.198 | 51.074 | 2.510 | 30.976 | 21.469 | 9.398 |
| $BVDF_{minimax}$ | 1.058 | 37.975 | 8.542 | 4.159 | 2.531 | 30.732 | 21.755 | 0.658 |
| $BVDF_{rgb}$ | 1.010 | 37.715 | 8.247 | 1.789 | 2.511 | 31.113 | 21.455 | 0.288 |

Note that the presented techniques benefit other directional filters including the adaptive nearest neighbor filter [14], hybrid directional filters [15], fuzzy *VDFs* [16], generalized *VDF* [17], adaptive *BVDF* [18], fuzzy hybrid filters [19], entropy filters [20], adaptive center weighted filters [10], and sigma vector filters [21].

## 3.2 Experiment 2

In order to further demonstrate the usefulness of the presented techniques, we evaluated their performance on a more recent filter, namely the *ACWDDF*. The formulation of this filter is given below:

$$\mathbf{y}(r,c) = \begin{cases} \mathbf{y}_{DDF} & if \ \sum_{k=\lambda}^{\lambda+2} A(\mathbf{y}_{CWDDF^k}, \mathbf{x}_d) \cdot L_p(\mathbf{y}_{CWDDF^k}, \mathbf{x}_d) > T \\ \mathbf{x}_d & otherwise \end{cases}, \quad \lambda \in [1, d-1]$$

$$\mathbf{y}_{CWVMF^k} = \underset{\mathbf{x}_i \in W}{argmin}\left( \sum_{j=1}^{n} w_j(k) \cdot L_p(\mathbf{x}_i, \mathbf{x}_j) \right) \qquad (9)$$

$$w_j(k) = \begin{cases} n-2k+2 & for \ \ j = d \\ 1 & otherwise \end{cases}, \quad k \in [1, d]$$

$$d = (n+1)/2$$

where $d$ is the index of the center pixel in $W$, $k$ is the smoothing parameter, $w_j(k)$ is the weight of pixel $\mathbf{x}_j$ at smoothing level $k$, $\lambda$ is a parameter that determines the initial smoothing level, $T$ is the switching threshold, $\mathbf{y}_{DDF}$ and $\mathbf{y}_{CWVMF^k}$ are the outputs of the *DDF* and center weighted *VMF* in $W$, respectively. The $\lambda$ and $T$ parameters were set to the author [10] recommended values of 2 and 10.8, respectively.

As mentioned in Section I, this filter was shown to be the most effective filter among 48 filters [3]. It was also shown that the *ACWDDF* is among the slowest (it ranked 41st). Therefore, this filter would benefit the most from the techniques presented in Section II.

The results of the first experiment showed that the characteristics of the function $A(\cdot, \cdot)$ can be captured very accurately using the function $B(\cdot, \cdot)$. This brings a question to mind: do functions $A(\cdot, \cdot)$ and $B(\cdot, \cdot)$ have a linear relationship? In order to test this, we generated $10^8$ random vector pairs in the *RGB* color space and calculated the distance between each pair using $A(\cdot, \cdot)$ and $B(\cdot, \cdot)$. We then calculated the best fitting line using the generalized least-squares method [22]:

$$A(\mathbf{x}_i, \mathbf{x}_j) \cong 1.436437 \cdot B(\mathbf{x}_i, \mathbf{x}_j) + 0.027664 \qquad (10)$$

The error of fit was $\varepsilon = 0.005715$. Based on the small error value it can be concluded that the relationship between $A(\cdot, \cdot)$ and $B(\cdot, \cdot)$ is in fact almost linear.

In the first experiment, we did not make use of (10) since in the formulation of *BVDF*, i.e. (2), all that is needed is an ordering function that behaves like $A(\cdot,\cdot)$. Since $A(\cdot,\cdot)$ and $B(\cdot,\cdot)$ have a highly linear relationship, the former function can be replaced with the latter without a significant loss of accuracy. However, in this experiment, we did not use $B(\cdot,\cdot)$ in its original form, i.e. (7). Instead, we plugged the distance values obtained using $B(\cdot,\cdot)$ into (10) to obtain an accurate approximation of $A(\cdot,\cdot)$.

Tables 4-5 show the filtering results for the original *ACWDDF* (denoted by *ACWDDF*), method 1 (denoted by *ACWDDF$_{minimax}$*), and method 2 (denoted by *ACWDDF$_{rgb}$*). As in the first experiment, in terms of filtering effectiveness, the *ACWDDF$_{minimax}$* consistently performed very similar to the *ACWDDF*. On the other hand, except for a few cases, the *ACWDDF$_{rgb}$* performed significantly better than the others. This might be due to the higher numerical stability of $B(\cdot,\cdot)$ when compared to $A(\cdot,\cdot)$. With respect to the execution speed, on the average, the *ACWDDF$_{minimax}$* and *ACWDDF$_{rgb}$* were respectively 6.67 and 13.49 times faster than the *ACWDDF*.

**Table 4.** *ACWDDF* filtering results at 10% noise level

| Filter | Airplane (512 x 512 pixels) | | | | Flowerbee (480 x 480 pixels) | | | |
|---|---|---|---|---|---|---|---|---|
| | MAE | PSNR | NCD | TIME | MAE | PSNR | NCD | TIME |
| none | 6.400 | 17.865 | 82.798 | 0.000 | 6.338 | 18.204 | 89.361 | 0.000 |
| *ACWDDF* | 0.646 | 35.005 | 6.673 | 11.302 | 0.441 | 38.615 | 4.049 | 9.577 |
| *ACWDDF$_{minimax}$* | 0.646 | 35.003 | 6.680 | 1.564 | 0.441 | 38.615 | 4.048 | 1.341 |
| *ACWDDF$_{rgb}$* | 0.587 | 35.955 | 5.896 | 0.775 | 0.439 | 38.807 | 3.883 | 0.669 |
| | Fractal (640 x 480 pixels) | | | | Lenna (512 x 512 pixels) | | | |
| none | 6.361 | 17.141 | 102.837 | 0.000 | 6.369 | 18.195 | 102.593 | 0.000 |
| *ACWDDF* | 1.002 | 30.831 | 19.021 | 9.884 | 0.544 | 38.483 | 6.986 | 11.347 |
| *ACWDDF$_{minimax}$* | 1.002 | 30.843 | 19.014 | 1.761 | 0.544 | 38.478 | 6.989 | 1.597 |
| *ACWDDF$_{rgb}$* | 1.065 | 31.027 | 20.187 | 0.875 | 0.536 | 38.801 | 6.734 | 0.773 |
| | Parrots (1536 x 768 pixels) | | | | Peppers (512 x 480 pixels) | | | |
| none | 6.348 | 18.123 | 117.100 | 0.000 | 6.371 | 17.964 | 95.057 | 0.000 |
| *ACWDDF* | 0.156 | 42.415 | 1.749 | 54.805 | 0.428 | 35.447 | 5.017 | 10.158 |
| *ACWDDF$_{minimax}$* | 0.156 | 42.414 | 1.749 | 8.956 | 0.428 | 35.461 | 5.014 | 1.430 |
| *ACWDDF$_{rgb}$* | 0.138 | 45.101 | 1.550 | 4.514 | 0.340 | 40.785 | 3.490 | 0.717 |

**Table 5.** *ACWDDF* filtering results at 15% noise level

| Filter | Airplane (512 x 512 pixels) | | | | Flowerbee (480 x 480 pixels) | | | |
|---|---|---|---|---|---|---|---|---|
| | MAE | PSNR | NCD | TIME | MAE | PSNR | NCD | TIME |
| none | 9.553 | 16.131 | 123.414 | 0.000 | 9.537 | 16.418 | 134.211 | 0.000 |
| *ACWDDF* | 1.029 | 32.691 | 11.044 | 11.280 | 0.688 | 35.905 | 6.520 | 9.516 |
| *ACWDDF$_{minimax}$* | 1.030 | 32.688 | 11.052 | 1.586 | 0.687 | 35.906 | 6.522 | 1.359 |
| *ACWDDF$_{rgb}$* | 0.951 | 33.347 | 9.940 | 0.773 | 0.677 | 36.235 | 6.190 | 0.672 |
| | Fractal (640 x 480 pixels) | | | | Lenna (512 x 512 pixels) | | | |
| None | 9.560 | 15.373 | 153.478 | 0.000 | 9.560 | 16.429 | 153.113 | 0.000 |
| *ACWDDF* | 1.325 | 29.381 | 24.604 | 9.842 | 0.853 | 35.656 | 11.137 | 11.266 |
| *ACWDDF$_{minimax}$* | 1.326 | 29.374 | 24.618 | 1.778 | 0.853 | 35.656 | 11.137 | 1.611 |
| *ACWDDF$_{rgb}$* | 1.334 | 30.023 | 24.774 | 0.875 | 0.835 | 35.995 | 10.693 | 0.775 |
| | Parrots (1536 x 768 pixels) | | | | Peppers (512 x 480 pixels) | | | |
| None | 9.514 | 16.363 | 175.817 | 0.000 | 9.576 | 16.199 | 142.114 | 0.000 |
| *ACWDDF* | 0.260 | 39.169 | 3.108 | 55.483 | 0.688 | 33.158 | 8.258 | 10.067 |
| *ACWDDF$_{minimax}$* | 0.260 | 39.151 | 3.112 | 9.113 | 0.688 | 33.158 | 8.258 | 1.450 |
| *ACWDDF$_{rgb}$* | 0.231 | 41.112 | 2.758 | 4.548 | 0.564 | 36.759 | 6.085 | 0.720 |

Figure 3 shows the filtering results for close-up parts of the *lenna* and *parrots* images. As expected, for both the *BVDF* and *ACWDDF* filters, methods 1 and 2 gave as good or better results when compared to the original *ACOS* implementations.

Note that the speed up that can be obtained by the use of the presented techniques in a particular filter depends on the contribution of $A(\cdot, \cdot)$ to the total filtering time. In the *BVDF*, $A(\cdot, \cdot)$ is the dominant factor in the computational time, whereas in the *ACWDDF* it has less influence on the total time. This is the reason why the computational gains were so different between the two experiments. Nevertheless, the results demonstrate that with the proposed modifications even the slowest directional filters can perform in real-time.

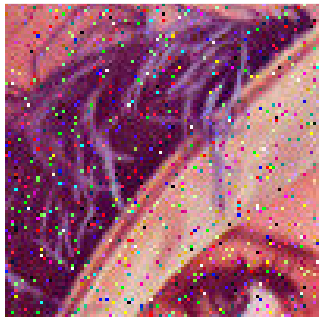

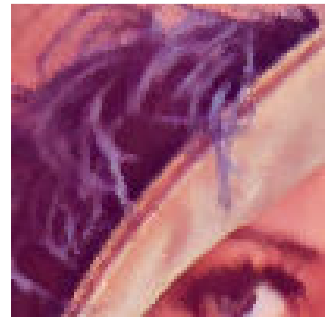

a) 10% noisy (MAE: 6.368681; PSNR: 18.194622)  b) BVDF (MAE: 3.929005; PSNR: 31.746791)

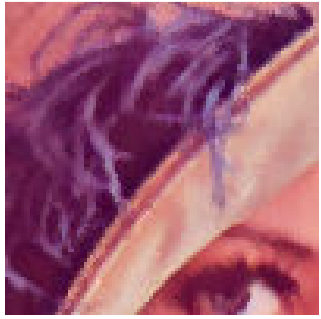 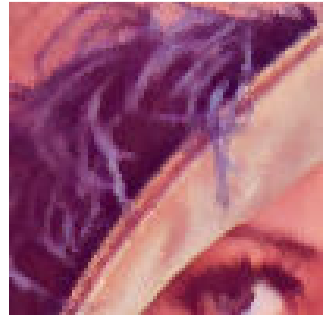

c) BVDF$_{minimax}$ (MAE: 3.929145; PSNR: 31.746964)   d) BVDF$_{rgb}$ (MAE: 3.932565; PSNR: 31.732303)

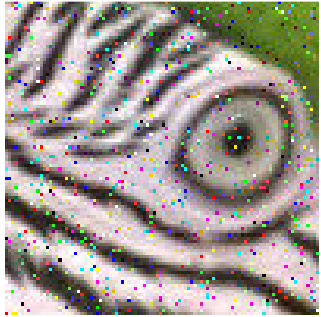 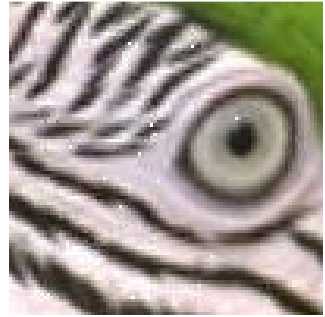

e) 10% noisy (MAE: 6.348155; PSNR: 18.123017)   f) ACWDDF (MAE: 0.156042; PSNR: 42.415066)

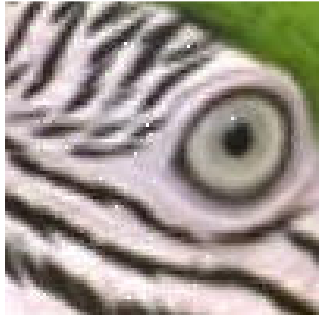 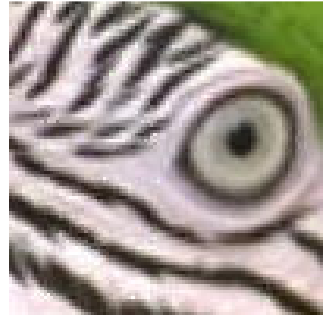

g) ACWDDF$_{minimax}$ (MAE: 0.156066; PSNR: 42.414146)   h) ACWDDF$_{rgb}$ (MAE: 0.138122; PSNR: 45.100753)

**Figure 3.** Filtering results for the *lenna* and *parrots* images corrupted with 10% noise

# IV. CONCLUSIONS

In this paper, we presented two methods to speed up order-statistics based directional filters. Experiments on a diverse set of color images showed that these methods can provide excellent accuracy and high computational gains. The presented approximation methods have applications that go beyond color image filtering including computer graphics and computational geometry.

The implementations of the filters described in this paper will be made publicly available as part of the Fourier image processing and analysis library, which can be downloaded from http://sourceforge.net/projects/fourier-ipal

# ACKNOWLEDGMENTS

This work was supported by a grant from the Louisiana Board of Regents (LEQSF2008-11-RD-A-12). The author is grateful to the anonymous reviewers for their valuable comments and suggestions and to Dr. Peter Alfeld (Department of Mathematics, The University of Utah) for providing the *fractal* image.